\documentclass[11pt,a4paper]{article}
\usepackage[hyperref]{emnlp2020}
\usepackage{times}
\usepackage{latexsym}

\usepackage{amsmath}
\usepackage{amssymb}
\usepackage{graphicx}
\usepackage[font=footnotesize]{caption}
\usepackage{url}
\usepackage{multirow}
\usepackage{floatrow}

\usepackage{microtype}

\aclfinalcopy 

\title{Hierarchical Encoders for Modeling and Interpreting Screenplays}

\author{Gayatri Bhat\thanks{\enskip Work done during an internship at Netflix} \\
  Bloomberg L.P. \\
  New York, NY, USA \\
  {\normalsize \texttt{gbhat7@bloomberg.net}} \\\And
  Avneesh Saluja \quad Melody Dye \quad Jan Florjanczyk \\
  Netflix \\
  Los Angeles, CA, USA \\
  {\normalsize \texttt{\{asaluja,mdye,jflorjanczyk\}@netflix.com}} \\}

\date{}

\begin{document}
\maketitle
\begin{abstract}
While natural language understanding of long-form documents is still an open challenge, such documents often contain structural information that can inform the design of models for encoding them.
Movie scripts are an example of such richly structured text – scripts are segmented into scenes, which are further decomposed into dialogue and descriptive components.
In this work, we propose a neural architecture for encoding this structure, which performs robustly on a pair of multi-label tag classification datasets, without the need for handcrafted features.
We add a layer of insight by augmenting an unsupervised ‘interpretability’ module to the encoder, allowing for the extraction and visualization of narrative trajectories.
Though this work specifically tackles screenplays, we discuss how the underlying approach can be generalized to a range of structured documents.
\end{abstract}

\section{Introduction}
As natural language understanding of sentences and short documents continues to improve, there has been growing interest in tackling longer-form documents such as academic papers \citep{Ren2014,Bhagavatula2018}, novels \citep{Iyyer2016} and screenplays \citep{Gorinski2018}. 
Analyses of such documents can take place at multiple levels, e.g. identifying both document-level labels (such as genre), as well as narrative trajectories (how do levels of humor and romance vary over the course of a romantic comedy?).
However, one of the key challenges for these tasks is that the signal-to-noise ratio over lengthy texts is generally low (as indicated by the performance of such models on curated datasets like NarrativeQA \citep{Kocisky2018}), making it difficult to apply end-to-end neural network solutions that have recently achieved state-of-the-art on other tasks \citep{Barrault2019,Williams2018,Wang2019}. 

Instead, models either rely on a) a \emph{pipeline} that provides a battery of syntactic and semantic information from which to craft features (e.g., the BookNLP pipeline \citep{Bamman2014} for literary text, graph-based features \citep{Gorinski2015} for movie scripts, or outputs from a discourse parser \citep{Ji2017} for text categorization) and/or b) the \emph{linguistic intuitions} of the model designer to select features relevant to the task at hand (e.g., rather than ingesting the entire text, \citet{Bhagavatula2018} only consider certain subsections like the title and abstract of an academic publication). 
While there is much to recommend these approaches, end-to-end neural modeling offers several key advantages: in particular, it obviates the need for auxiliary feature-generating models, minimizes the risk of error propagation, and offers improved generalization across large-scale corpora.
This work explores how models can leverage the inherent structure of a document class to facilitate an end-to-end approach.
Here, we focus on screenplays, investigating whether we can effectively extract key information by first segmenting them into scenes, and then further exploiting the structural regularities within each scene.

With an average of \textgreater 20k tokens per script in our evaluation corpus, extracting salient aspects is far from trivial. 
Through a series of carefully controlled experiments, we show that a structure-aware approach significantly improves document classification by effectively collating sparsely distributed information. 
Further, this method produces both document- and scene-level embeddings, which can be used downstream to visualize narrative trajectories of interest (e.g., the prominence of various themes across the script).
The overarching strategy of this work is to incorporate structural priors as biases into the \emph{architecture} of the neural network model itself (e.g., \citet{Socher2013}, \citet{Strubell2018}, \emph{inter alia}).
The methods we propose can readily generalize to any long-form text with an exploitable internal structure, including novels (chapters), theatrical plays (scenes), chat logs (turn-taking), online games (levels/rounds/gameplay events), and academic texts (sections and subsections).

The paper is organized as follows: In \S\ref{sec:script_structure}, we detail how a script can be formally decomposed into scenes, and each scene can be further decomposed into granular elements with distinct discourse functions. \S\ref{sec:hse} elaborates on how this structure can be effectively leveraged with a proposed encoder based on hierarchical attention \citep{Yang2016}. In \S\ref{sec:tag_prediction}, the predictive performance of the hierarchical encoder is validated on two multi-label tag prediction tasks, one of which rigorously establishes the utility of modeling structure at multiple granularities (i.e., at the level of line, scene, and script).
Notably, while the resulting scene-encoded representation is useful for prediction tasks, it is not amenable to easy interpretation or examination. 
To shed further light on encoded document representation, in \S\ref{sec:unsup_dict_learn}, we propose an unsupervised interpretability module that can be attached to an encoder of any complexity. 
\S\ref{sec:qual_analysis} outlines our application of this module to the scene encoder, and the resulting visualizations of the screenplay, which neatly illustrate how plot elements vary over the course of the narrative arc. 
\S\ref{sec:related} draws connections to related work, before concluding.

\section{Script Structure}
\label{sec:script_structure}
\begin{figure}[!t]
    \begin{center}
        \includegraphics[height=0.15\textheight, width=0.9\textwidth]{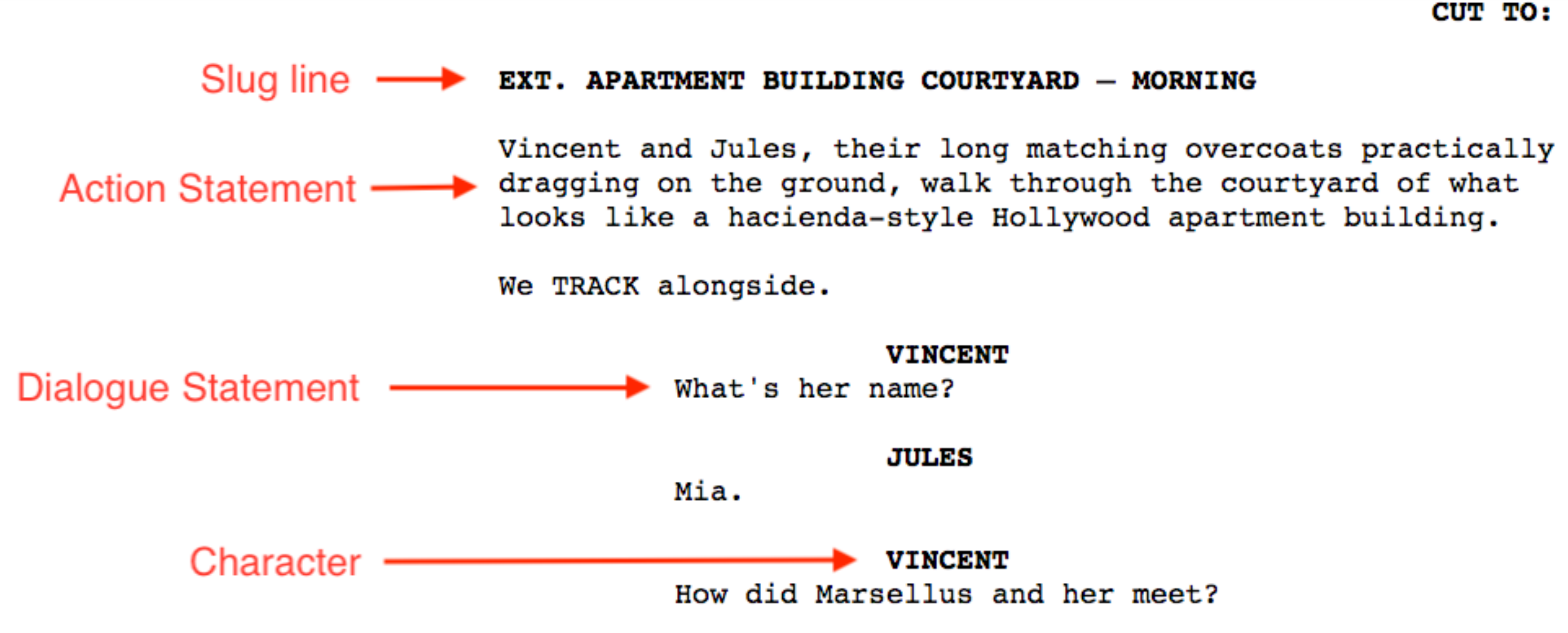}
    \end{center}
    \caption{A portion of the screenplay for \emph{Pulp Fiction}, annotated with the common scene components.}
    \vspace{-0.3cm}
    \label{fig:example_screenplay}
\end{figure}
Movie and television scripts, also known as screenplays, are traditionally segmented into \emph{scenes}, with a rough rule of thumb that each scene lasts about a minute on-screen. 
A scene is not necessarily a distinct narrative unit (which are most often sequences of several consecutive scenes), but is constituted by a piece of continuous action at a single location.
\begin{table}[!h]
    \begin{center}
    \scriptsize
    \begin{tabular}{p{1.2cm}p{12pt}p{12pt}p{0.5cm}p{1cm}p{1.8cm}}
        Title& Line& Scene& Type& Character& Text \\\hline
        Pulp Fiction& 204& 4& Scene & & {EXT. APART..} \\
        Pulp Fiction& 205& 4& Action & & {Vincent and Jules.} \\
        Pulp Fiction& 206& 4& Action & & {We TRACK...} \\
        Pulp Fiction& 207& 4& Dial. & VINCENT & What's her name?\\
        Pulp Fiction& 208& 4& Dial. & JULES & Mia. \\
        Pulp Fiction& 209& 4& Dial. & VINCENT & How did... \\
    \end{tabular}
    \end{center}
    \vspace{-0.3cm}
\caption{Post-processed version of Fig.\ref{fig:example_screenplay}.}
\label{tab:processed_script}
\end{table}

Fig. \ref{fig:example_screenplay} contains a segment of a scene from the screenplay for the movie \emph{Pulp Fiction}, a 1994 American film.
These segments tend to follow a standard format.
Each scene starts with a scene heading or ``slug line" that briefly describes the scene setting, followed by a sequence of statements.
Screenwriters typically use formatting to distinguish between dialogue and action statements \citep{Argentini1998}.
The first kind contains lines of a dialogue and identifies the character who utters it either on- or off-screen (the latter is often indicated with `(V.O.)' for voice-over).
Occasionally, parentheticals are used to include special instructions for how an utterance should be delivered by the character.
Action statements, on the other hand, are all non-dialogue constituents of the screenplay ``often used by the screenwriter to describe character actions, camera movement, appearance, and other details" \citep{Pavel2015}. 
In this work, we consider action and dialogue statements, as well as character identities for each dialogue segment, and ignore slug lines and parentheticals. 

\section{Hierarchical Scene Encoders}
\label{sec:hse}
\begin{figure}[!htbp]
    \begin{center}
        \includegraphics[width=0.9\textwidth,keepaspectratio,trim={0 3.5cm 11.5cm 0.5cm}]{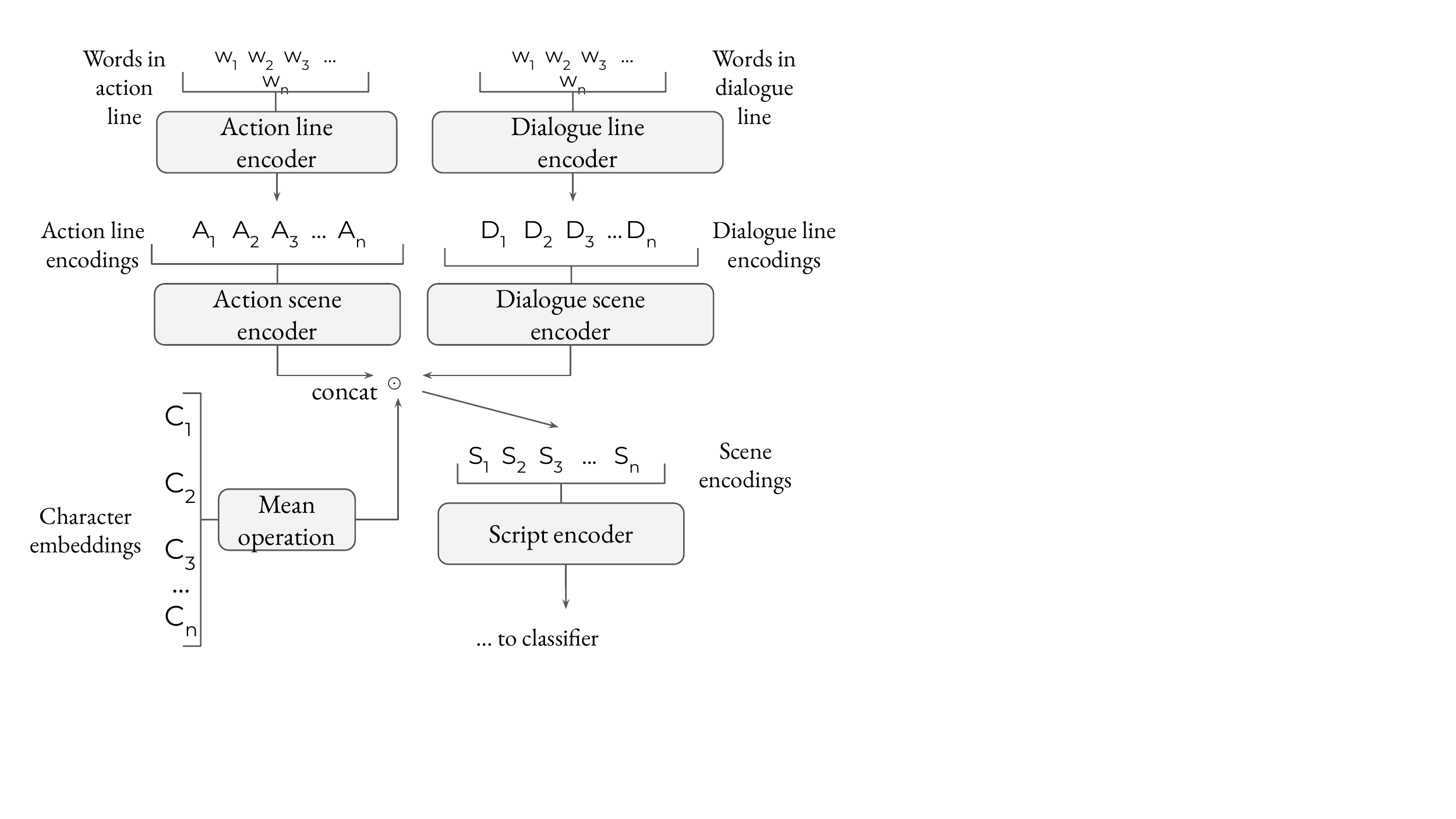}
    \end{center}
    \caption{The architecture of our script encoder, largely following the structure in Fig. \ref{fig:example_screenplay}.}
    \label{fig:model_schematic}
\end{figure}

Given the size of a movie script, it is computationally infeasible to treat these screenplays as single blocks of text to be ingested by a recurrent encoder. 
Instead, we propose a hierarchical encoder that mirrors the standard structure of a screenplay (\S\ref{sec:script_structure}) -- a sequence of scenes, each of which is in turn an interwoven sequence of action and dialogue statements. 
The encoder is three-tiered, as illustrated in Fig. \ref{fig:model_schematic} and processes the text of a script as follows. 

\subsection{Model Architecture}
\label{sec:model_architecture}
First, an \textbf{action-statement encoder} transforms the sequence of words in an action statement (represented by their pretrained word embeddings) into an action statement embedding. 
Next, an \textbf{action-scene encoder} transforms the chronological sequence of action statement embeddings within a scene into an action scene embedding. 
Analogously, a \textbf{dialogue-statement encoder} and a \textbf{dialogue-scene encoder} are used to obtain dialogue statement embeddings and aggregate them into dialogue scene embeddings.
To evaluate the effect of character information, characters with at least one dialogue statement in a given scene are represented by an individual character embedding (these are randomly initialized and estimated during model training), and a scene-level character embedding is constructed by averaging the embeddings of all the characters in the scene\footnote{We only take into account characters at the \emph{scene} level i.e., we do not associate characters with each dialogue statement, leaving this addition to future work.}.
Finally, the action, dialogue and scene-level character embeddings for each scene are concatenated into a single scene embedding.
\vspace{-0.25cm}

Scene-level predictions or analyses can then be obtained by feeding the scene embeddings into a subsequent module of the neural architecture, e.g. a feedforward layer can be used for supervised tagging tasks.
Alternatively, if a single representation of the entire screenplay is required, a final \textbf{script encoder} is used to transform the sequence of scene embeddings for a script into a single script embedding.
A key assumption underlying the model is that action and dialogue statements -- as instances of written narrative and spoken language respectively -- are distinct categories of text and must therefore be processed separately.
We evaluate this assumption in the tag classification experiments (\S\ref{sec:tag_prediction}).

\subsection{Encoders}
\label{sec:encoders}
The proposed model incorporates strong inductive biases regarding the overall structure of input documents.
In addition, each of the aforementioned encoders in \S\ref{sec:model_architecture} can be specified in multiple ways, and we evaluate three different instantiations of the encoder components:
\begin{enumerate}
    \item \textbf{Sequential (GRU):} A bidirectional GRU \citep{Bahdanau2015} is used to encode the temporal sequence of inputs (of words, statements or scenes). 
    Given a sequence of input embeddings $\mathbf{e}_1, \dots, \mathbf{e}_T$ for a sequence of length $T$, we obtain GRU outputs $\mathbf{c}_1, \dots, \mathbf{c}_T$, and use $\mathbf{c}_T$ as the recurrent encoder's final output.
    Other sequential encoders could also be used as alternatives.
    \item \textbf{Sequential with Attention (GRU + Attn):} Attention \citep{Bahdanau2015} can be used to combine sequential outputs $\mathbf{c}_1, \dots, \mathbf{c}_T$, providing a mechanism for more or less informative inputs to be filtered accordingly.
    We calculate attention weights using a parametrized vector $\mathbf{p}$ of the same dimensionality as the GRU outputs \citep{Sukhbaatar2015,Yang2016}:
    \begin{align*}
        \alpha_i = \frac{\mathbf{p}^T \mathbf{c}_i}{\Sigma_{j=1}^{T}\mathbf{p}^T \mathbf{c}_j}
    \end{align*}
    These weights are used to compute the final output of the encoder as:
    \begin{align*}
        \mathbf{c} = \Sigma_{j=1}^{T} \alpha_i \mathbf{c}_i
    \end{align*}
    Other encoders with attention could be used as alternatives to this formulation.
    \item \textbf{Bag-of-Embeddings with Attention (BoE + Attn):} Another option is to disregard the sequential encoding and simply compute an attention-weighted average of the inputs to the encoder as follows:
    \begin{align*}
        \alpha_i &= \frac{\mathbf{p}^T \mathbf{e}_i}{\Sigma_{j=1}^{T}\mathbf{p}^T \mathbf{e}_j}\\
         \mathbf{c} &= \Sigma_{j=1}^{T} \alpha_i \mathbf{e}_i
    \end{align*}
    This encoder stands in contrast to a bag-of-embeddings (BoE) encoder which computes a simple average of its inputs. 
    While defining a far more constrained function space than recurrent encoders, BoE and BoE + Attn representations have the advantage of being interpretable (in the sense that the encoder's output is in the same space as the input word embeddings).
    We leverage this property in \S\ref{sec:unsup_dict_learn} where we develop an interpretability layer on top of the encoder outputs.
\end{enumerate}

\subsection{Loss for Tag Classification}
\label{sec:classifier_loss}
The final script embedding being passed into a feedforward classifier (FFNN).
As both supervised learning tasks in our evaluation are multi-label classification problems, we use a variant of a simple multi-label one-versus-rest loss, where correlations among tags are ignored.
The tag sets have high cardinalities and the fractions of positive samples are inconsistent across tags (Table \ref{tab:tag_cardinality} in \ref{sec:add_stats}); this motivates us to train the model with a reweighted loss function:
\begin{align}
L(y,z) &= \tfrac{1}{NL} \Sigma_{i=1}^{N} \Sigma_{j=1}^{L} y_{ij} \log\sigma(z_{ij}) \nonumber \\
    &+ \lambda_j (1 - y_{ij}) (1 - \log\sigma(z_{ij})) \label{eq:loss_fn}
\end{align}
where $N$ is the number of samples, $L$ is the number of tag labels, $y \in \{0, 1\}$ is the tag label, $z$ is the output of the FFNN, $\sigma$ is the sigmoid function, and $\lambda_j$ is the ratio of positive to negative samples (precomputed over the entire training set, since the development set is too small to tune this parameter) for the tag label indexed by $j$. 
With this loss function, we account for label imbalance without using separate thresholds for each tag tuned on the validation set.

\section{Interpreting Scene Embeddings}
\label{sec:unsup_dict_learn}
As the complexity of learning methods used to encode sentences and documents has increased, so has the need to understand the properties of the encoded representations.
Probing-based methods \citep{Linzen2016, Conneau2018} are used to gauge the information captured in an embedding by evaluating its performance on downstream classification tasks, either with manually collected annotations \citep{Shi2016} or carefully selected self-supervised proxies \citep{Adi2016}.
In our case, it is laborious and expensive to collect such annotations at the scene level (requiring domain experts), and the proxy evaluation tasks proposed in the literature do not probe the narrative properties we wish to surface.

Instead, we take inspiration from \citet{Iyyer2016} and learn a \textbf{scene descriptor model} that can be trained without relying on any such annotations.
Using a dictionary learning perspective \citep{Olshausen1997}, the model learns to represent each scene embedding as a weighted mixture of various topics estimated over the entire corpus.
It thus acts as an ``interpretability layer" that can be applied over the scene encoder.
This model class is similar in spirit to dynamic topic models \citep{Blei2006}, with the added advantage of producing topics that are both more coherent and more interpretable than those generated by LDA \citep{He2017unsupervised,Mitcheltree2018}.

\subsection{Scene Descriptor Model}
The model has three main components: a \textbf{scene encoder}, a set of topics or \textbf{descriptors} that form the ``basis elements" used to describe an interpretable scene, and a \textbf{predictor} that predicts weights over descriptors for a given scene embedding. 
The scene encoder uses the text of a given scene $s_t$ to produce a corresponding scene embedding $\mathbf{v}_t$.
This encoder can take any form -- from an extractor that derives a hand-crafted feature set from the scene text, as in \citet{Gorinski2018}, to an instantiation of the scene encoder in \S\ref{sec:hse}. 

\vspace{-0.25cm}
\begin{figure}[!htbp]
    \begin{center}
        \includegraphics[width=0.75\textwidth,keepaspectratio,trim={1cm 0 6cm 0}]{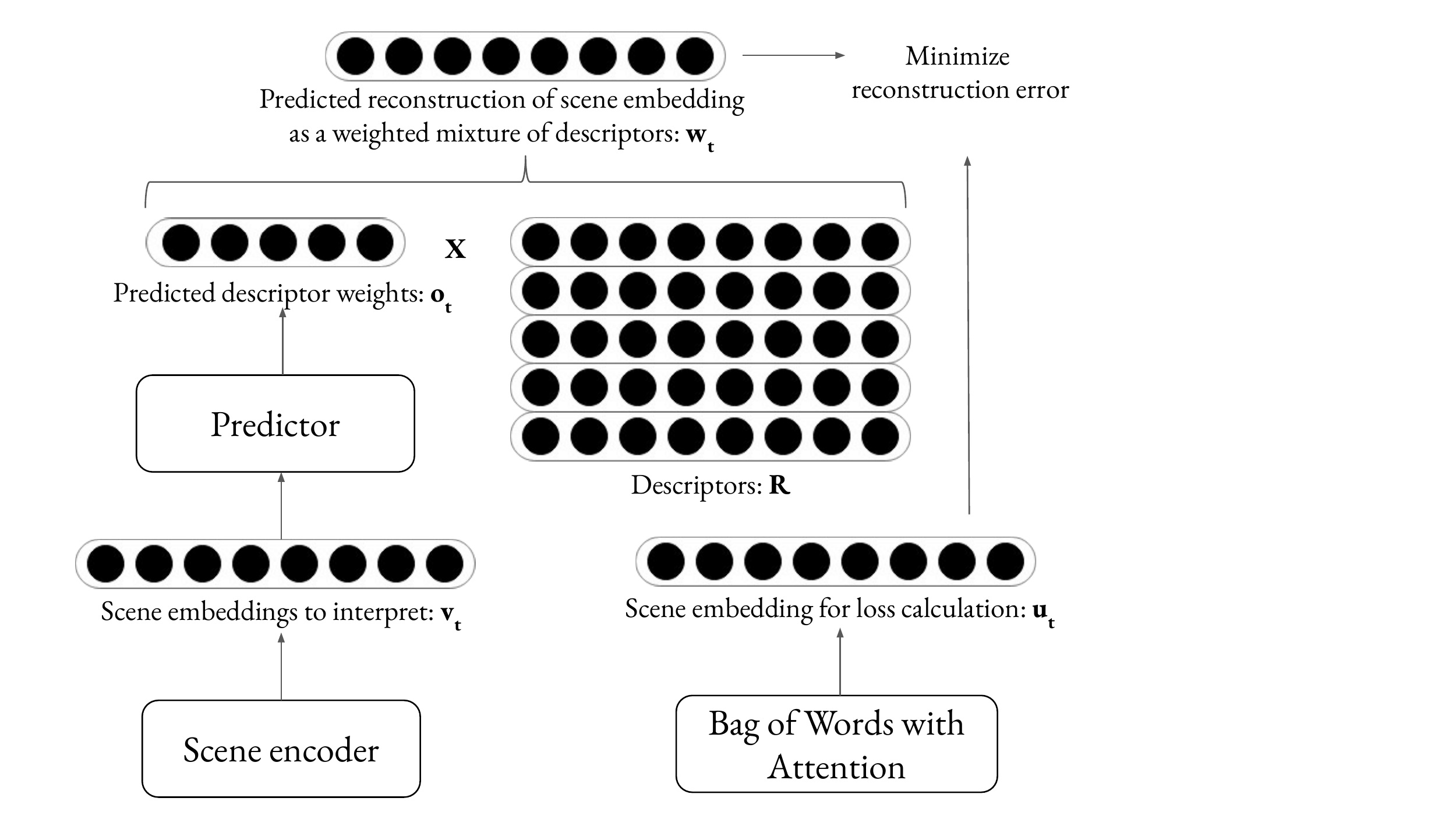}
    \end{center}
    \caption{A pictorial representation of the scene descriptor model.}
    \label{fig:descriptor_schematic}
\end{figure}
\vspace{-0.3cm}
To probe the contents of scene embedding $\mathbf{v}_t$, we compute the descriptor-based representation $\mathbf{w}_t \in \mathbb{R}^d$ in terms of a descriptor matrix $\mathbf{R}^{k \times d}$, where $k$ is the number of topics or descriptors:
\begin{align}
    \mathbf{o}_t &= \textrm{softmax}(f(\mathbf{v}_t)) \label{eq:predictor} \\
    \mathbf{w}_t &= \mathbf{R}^T \mathbf{o}_t \nonumber
\end{align}
where $\mathbf{o}_t \in \mathbb{R}^k$ is the weight (probability) vector over $k$ descriptors and $f(\mathbf{v}_t)$ is a predictor (illustrated by the leftmost pipeline in  Fig. \ref{fig:descriptor_schematic}) which converts $\mathbf{v}_t$ into $\mathbf{o}_t$.
Two variants are $f = \textrm{FFNN}(\mathbf{v}_t)$ and $f = \textrm{FFNN}([\mathbf{v}_t; \mathbf{o}_{t-1}])$ (concatenation); we use the former in \S\ref{sec:qual_analysis}.
Furthermore, we can incorporate additional recurrence into the model by modifying Eq. \ref{eq:predictor} to add the previous state:
\begin{align}
    \mathbf{o}_t = (1-\alpha)\cdot\textrm{FFNN}([\mathbf{v}_t; \mathbf{o}_{t-1}]) + \alpha\cdot\mathbf{o}_{t-1} \label{eq:recurrent_predictor}
\end{align}

\subsection{Reconstruction Task}
We wish to minimize the reconstruction error between two scene representations: (1) the descriptor-based embedding $\mathbf{w}_t$ which depends on the scene embedding $\mathbf{v}_t$, and (2) an attention-weighted bag-of-words embedding for $s_t$.
This ensures that the computed descriptor weights are indicative of the scene's actual content (specifically portions of its text that indicate attributes of interest such as genre, plot, and mood).
We use a \texttt{BoE+Attn} scene encoder (\S\ref{sec:encoders}) pretrained on the tag classification task (bottom right of Fig. \ref{fig:descriptor_schematic}), which yields a vector $\mathbf{u}_t \in \mathbb{R}^d$ for scene $s_t$.
The scene descriptor model is then trained using a hinge loss objective \citep{Weston2011} to minimize the reconstruction error  between $\mathbf{w_t}$ and $\mathbf{u}_t$, with an additional orthogonality constraint on $\mathbf{R}$ to encourage semantically distinct descriptors:
\begin{equation}
   L = \Sigma_{j=1}^{n} \max(0, 1 - \mathbf{w}_t^T \mathbf{u}_t + \mathbf{w}_t^T \mathbf{u}_j ) + \lambda \| \mathbf{RR}^T - \mathbf{I} \|_2
\end{equation}
where $\mathbf{u}_1 \dots \mathbf{u}_n$ are $n$ negative samples selected from other scenes in the same screenplay.

The motivation for using the output of a \texttt{BoE+Attn} scene encoder is that $\mathbf{w}_t$ (and therefore the rows in $\mathbf{R}$) lies in the same space as the input word embeddings.
Thus, a given descriptor can be semantically interpreted by querying in the word embedding space
The predicted descriptor weights for a scene $s_t$ can be obtained by running a forward pass through the model.

\section{Evaluation}
\label{sec:evaluation}
We evaluate the proposed script encoder architecture and its variants through two supervised multi-label tag prediction tasks, and a qualitative analysis based on extracting descriptor trajectories in an unsupervised setting.

\subsection{Datasets}
\label{sec:datasets}
Our evaluation is based on the ScriptBase-J corpus, released by \citet{Gorinski2018}.\footnote{https://github.com/EdinburghNLP/scriptbase}
In this corpus, each movie is associated with a set of expert-curated tags that range across 6 tag attributes: mood, plot, genre, attitude, place, and flag (\ref{tab:tag_examples}); in addition to evaluating on these tags, we also used an internal dataset, where the same movies were hand-labeled by in-house domain experts across 3 tag attributes: genre, plot, and mood.
The tag taxonomies between these two datasets are distinct (Table \ref{tab:tag_cardinality}). 
ScriptBase-J was used both to directly compare our approach with an implementation of the multilabel encoder architecture in \citet{Gorinski2018} and to provide an open-source evaluation standard.

\subsubsection*{Script Preprocessing}
As in \citet{Pavel2015}, we leveraged the standard screenplay format \citep{Argentini1998} to extract a structured representation of the scripts (relevant formatting cues included capitalization and tab-spacing; see Fig. \ref{fig:example_screenplay} and Table \ref{tab:processed_script} for an example). 
Filtering erroneously processed scripts removed 6\% of the corpus, resulting in 857 scripts total. 
We set aside 20\% (172 scripts) for heldout evaluation; the remainder was used for training. 
The average number of tokens per script is around 23k; additional statistics are shown in Table \ref{tab:token_stats}.

Next, we split extremely long scenes into smaller ones, capping the maximum number of lines in a scene (across both action and dialogue) to 60 (keeping within GPU memory limits). 
For the vocabulary, a word count of 5 across the script corpus was set as the minimum threshold. 
The number of samples (scripts) per tag value ranges from high (e.g., for some genre tags) to low (for most plot and mood tags) in both datasets (\S\ref{sec:add_stats}), and coupled with high tag cardinality for each attribute, motivates the need for the reweighted loss in Eq.~\ref{eq:loss_fn}.

\subsection{Experimental Setup}
All inputs to the hierarchical scene encoder are 100-dimensional GloVe embeddings \cite{Pennington2014}.\footnote{Using richer contextual word representations will improve performance, but is orthogonal to the purpose of this work.}
Our sequential models are  biGRUs with a single 50-dimensional hidden layer in each direction, resulting in 100-dimensional outputs.
The attention model is parametrized by a 100-dimensional vector $\mathbf{p}$; BoE models naturally output 100-dimensional representations, and character embeddings are 10-dimensional.
The output of the script encoder is passed through a linear layer with a sigmoid activation function and binarized by thresholding at 0.5.

One simplification in our experiments is to utilize the same encoder \emph{type} for all encoders described in \S\ref{sec:model_architecture}.
However, it is conceivable that different encoder types might perform better at different tiers of the architecture: e.g. scene aggregation can be done in a permutation-invariant manner, since narratives are interwoven and scenes may not be truly sequential.

We implement the script encoder on top of AllenNLP \citep{Gardner2017} and PyTorch \citep{Paszke2019}, and all experiments were conducted on an AWS \texttt{p2.8xlarge} machine.
We use the Adam optimizer with an initial learning rate of $5e^{-3}$, clip gradients at a maximum norm of 5, and do not use dropout.
The model is trained for a maximum of 20 epochs to maximize average precision score, and with early stopping in place if the validation metric does not improve for 5 epochs.

\subsection{Tag Prediction Experiments}
\label{sec:tag_prediction}
ScriptBase-J also comes with ``loglines", or short, 1-2 sentence human-crafted summaries of the movie's plot and mood (see Table \ref{tab:tag_examples}).
A model trained on these summaries can be expected to provide a reasonable baseline for tag prediction, since human summarization is likely to pick out relevant parts of the text for this task.
The loglines model is a bidirectional GRU with inputs of size 100 (GloVe embeddings) and hidden units of size 50 in each direction, whose output feeds into a linear classifier.\footnote{We tried both with and without attention and found the variant without attention to give slightly better results.}
\begin{table}[!h]
\vspace{-0.4cm}
\begin{tabular}{lrrr}
{\textbf{Model}} & {\textbf{Genre}} & {\textbf{Plot}} & {\textbf{Mood}} \\\hline
Loglines & 49.9 (0.8) & 12.7 (0.9) & 17.5 (0.2) \\
\hline
\multicolumn{4}{l}{\textit{Comparing encoder variations:}}\\
BoE & 49.0 (1.1) & 8.3 (0.6) & 12.9 (0.7) \\
BoE + Attn & 51.9 (2.3) & 11.3  (0.4) & 16.3 (0.6) \\
GRU & 57.9 (1.9) & 13.0 (1.3) & 19.1 (1.0) \\
GRU + Attn & 60.5 (2.0) & {\bf 15.2} (0.4) & {\bf 22.9} (1.4) \\
\hline
\multicolumn{4}{l}{\textit{Variants on GRU + Attn for action \& dialog:}} \\
+ Chars & {\bf 62.5} (0.7) & 11.7 (0.3) & 18.2 (0.3) \\
- Action & 60.5 (2.9) & 13.5 (1.4) & 20.0 (1.2) \\
- Dialogue & 60.5 (0.6) & 13.4 (1.7) & 19.1 (1.4) \\
2-tier & 61.3 (2.3) & 13.7 (1.7) & 20.6 (1.2) \\
HAN & 61.5 (0.6) & 14.2 (1.7) & 20.7 (1.4)
\end{tabular}
\caption{Investigation of the effects of different architectural (BoE +/- Attn, GRU +/- Attn) and structural choices on a tag prediction task, using an internally tagged dataset: F-1 scores with sample standard deviation in parentheses. Across the 3 tag attributes we find that modeling sentential and scene-level structure helps, and attention helps extract representations more salient to the task at hand.}
\label{tab:content_exps}
\vspace{-0.2cm}
\end{table}

Table \ref{tab:content_exps} contains results for the tag prediction task on our internally-tagged dataset.
First, a set of models trained using action and dialogue inputs are used to evaluate the architectural choices in \S\ref{sec:model_architecture}.
We find that modeling recurrence at the sentential and scene levels, and using attention to select relevant words or scenes, help considerably and are necessary for robust improvement over the loglines' baseline (see the first five rows in Table \ref{tab:content_exps}). 

Next, we assess the effect that various structural elements of a screenplay have on classification performance. Notably, the difficulty of the prediction task is directly related to the set size of the tag attribute: higher-cardinality tag attributes with correlated tag values (like plot and mood)  are significantly more difficult to predict than lower-cardinality tags with more discriminable values (like genre).
We find that adding character information to the best-performing GRU + Attn model (\texttt{+Char}) improves prediction of genre, while using both dialogue and action statements improves performance on plot and mood, compared to using only one or the other.
We also evaluate (1) a \texttt{2-tier} variant of the \texttt{GRU+Attn} model without action/dialogue-statement encoders (i.e., all action statements are concatenated into a single sequence of words and passed into the action-scene encoder, and similarly with dialogue) and (2) a variant similar to \citet{Yang2016} (\texttt{HAN}) that does not distinguish between action and dialogue (i.e., all statements in the text of a scene are encoded using a statement encoder and statement embeddings are passed to a single scene encoder, the output of which is passed into the script encoder).
Both models perform slightly better than \texttt{GRU+Attn} on genre, but worse on plot and mood, showing that for more difficult prediction tasks, it helps to incorporate hierarchy and to distinguish action and dialogue statements.

\begin{table}[!h]
\vspace{-0.2cm}
\begin{center}
\begin{tabular}{l|rr}
\textbf{Tag} & \textbf{G\&L} & \textbf{HSE} \\
\hline 
Attitude & 72.6 & 70.1 \\
Flag & 52.5 & 52.6 \\
Genre & 55.1 & 42.5 \\
Mood & 45.5 & 51.2 \\
Place & 57.7 & 29.1 \\
Plot & 34.6 & 34.5
\end{tabular}
\end{center}
\vspace{-0.4cm}
\caption{F-1 scores on ScriptBase-J provided tag set, comparing \citet{Gorinski2018}'s approach to ours.}
\label{tab:jinni}
\end{table}
For the results in Table \ref{tab:jinni}, we compared the \texttt{GRU+Attn} configuration in Table \ref{tab:content_exps} (\texttt{HSE}) with an implementation of \citet{Gorinski2018} (\texttt{G\&L}) that was run on the previous train-test split.
\texttt{G\&L} contains a number of handcrafted lexical, graph-based, and interactive features that were designed for optimal performance for screenplay analysis. 
In contrast, \texttt{HSE} directly encodes standard screenplay structure into a neural network architecture, and is an alternative, arguably more lightweight way of building a domain-specific textual representation.
Our results are comparable, with the exception of ``place", which can often be identified deterministically from scene headings.

\subsection{Similarity-based F-1}
\label{sec:similarity_scoring}
Results in Tables \ref{tab:content_exps} and \ref{tab:jinni} are stated using standard multi-label F-1 score (one-vs-rest classification evaluation, micro-averaged over each tag attribute), which requires an exact match between predicted and actual tag value to be deemed correct.
However, the characteristics of our tag taxonomies suggest that this measure may not be ideal. 
In particular, our human-crafted tag sets have tag attributes with dozens of highly correlated, overlapping values, as well as missing tags not assigned by the annotator.
A standard scoring procedure may underestimate model performance when, e.g., a prediction of ``Crime" for a target label of ``Heist", is counted as equivalently wrong to ``Romance" (Table \ref{tab:tag_similarity} in \ref{sec:add_stats}). 

One way to deal with tag sets is to leverage a similarity-based scoring procedure (see \citet{Maynard2006} for related approaches).
Such a measure takes into account the latent relationships among tags via \emph{similarity thresholding}, wherein a prediction is counted as correct if it is within a certain distance of the target.
In particular, we treat a prediction as correct based the percentile of its similarity to the actual label. 
The percentile cutoff can be varied to illustrate how estimated model performance varies as a function of the degree of ``enforced" similarity between target and prediction.

\begin{figure}[t!]
\centering
    \includegraphics[width=0.9\textwidth,keepaspectratio,trim={1cm, 1cm 0cm 0cm}]{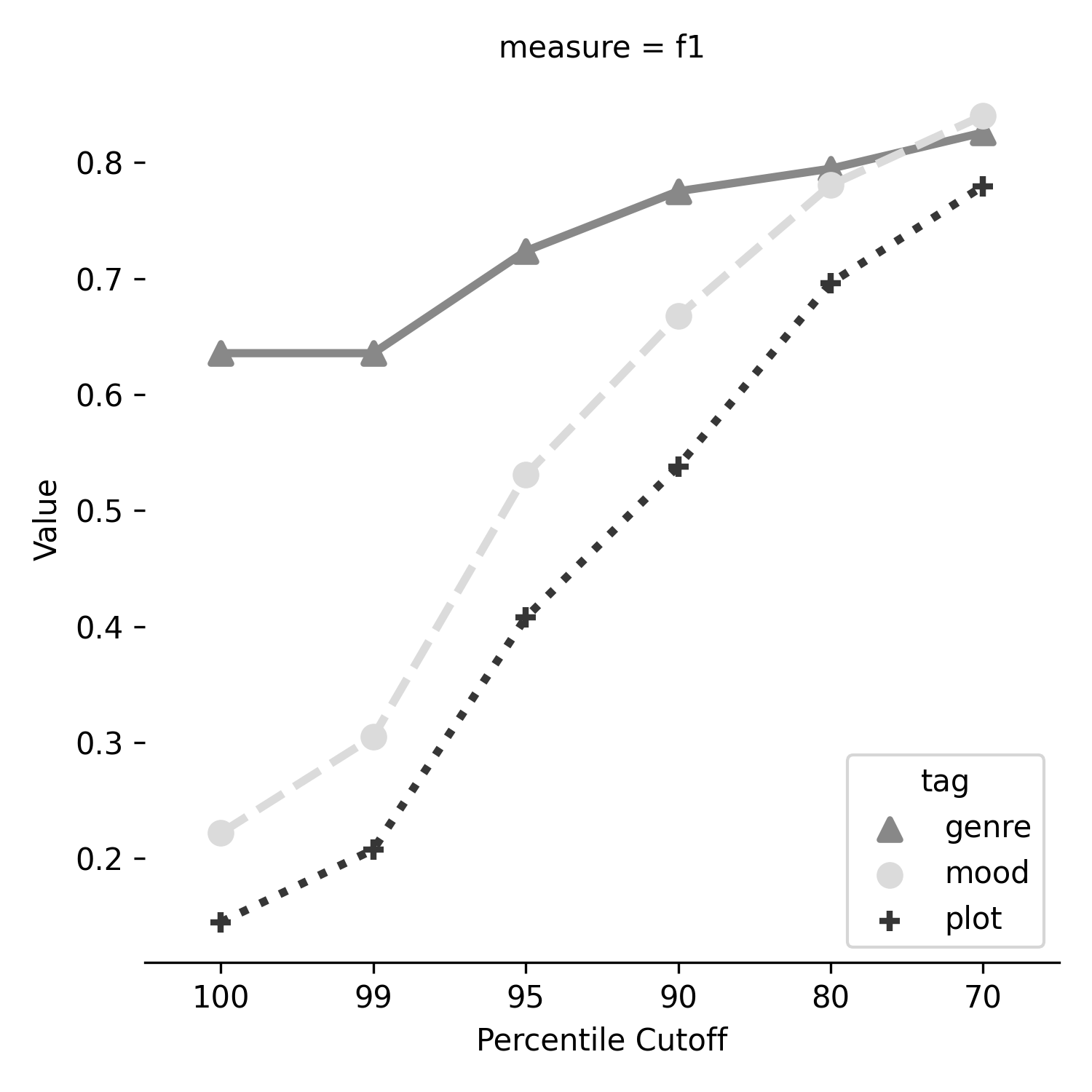}
\caption{F1 score of various tag attributes as a function of the similarity threshold percentile.}
\label{fig:sim_improvements}
\end{figure}
In Fig.~\ref{fig:sim_improvements} we examine how our results might vary if we adopted a similarity-based scoring procedure, by re-evaluating the \texttt{GRU + Attn} model outputs (row 5 in Table \ref{tab:content_exps}) with this evaluation metric.
When the similarity percentile cutoff equals 100, the result is identical to the standard F-1 score.
Even decreasing the cutoff to the 90\textsuperscript{th} percentile shows striking improvements for high-cardinality attributes (180\% for \texttt{mood} and 250\% for \texttt{plot}).
Leveraging a similarity-based scoring procedure for complex tag taxonomies may yield results that more accurately reflect human perception of the model's performance \citep{Maynard2006}.

\subsection{Qualitative Scene-level Analysis}
\label{sec:qual_analysis}
\begin{figure*}[t!]
\centering
    \includegraphics[width=0.75\paperwidth,keepaspectratio]{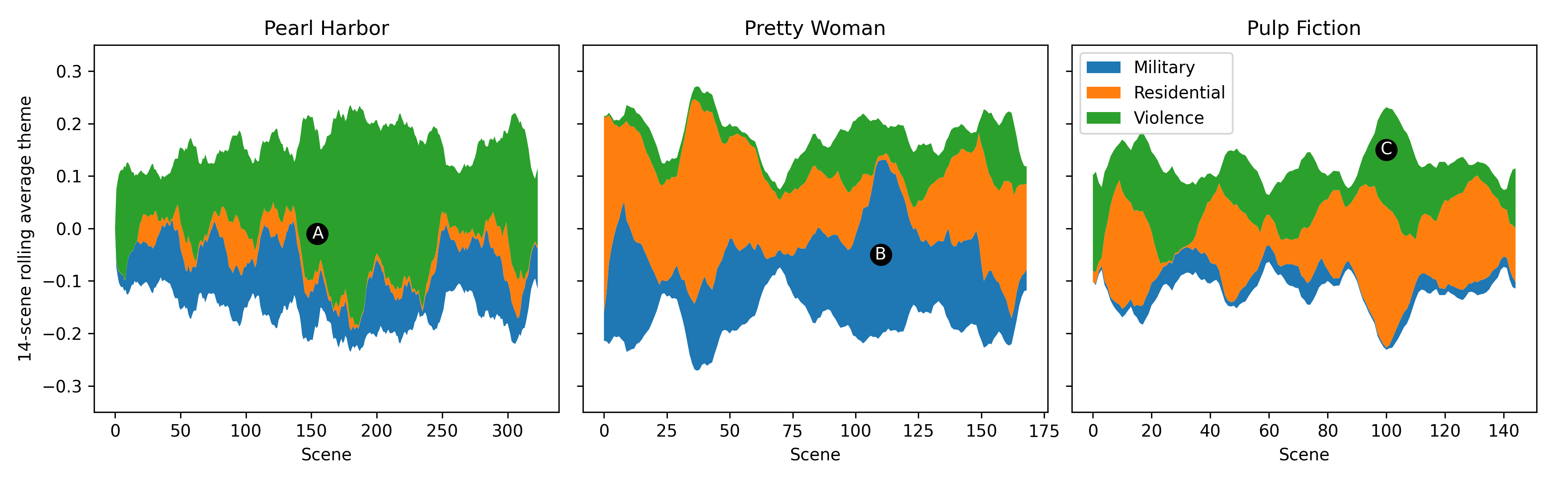}
\caption{Descriptor Trajectories for \emph{Pearl Harbor}, \emph{Pretty Woman}, and \emph{Pulp Fiction}. The $y$-axis is a smoothed and rescaled descriptor weight, i.e. $\mathbf{o}_t$ in Eq.\ref{eq:predictor}. Events: (A) Attack on Pearl Harbor begins (B) Rising tension at the equestrian club and (C) Confrontation at the pawn shop. Word clusters corresponding to each descriptor are in Table \ref{tab:clusters}.}
\label{fig:descriptor_trajectories}
\end{figure*}
To extract narrative trajectories with the scene descriptor model, we compared the three model variants in \S\ref{sec:model_architecture} for the choice of scene encoder and found that while attention aids the creation of interpretable descriptors (in-line with previous work), sequential and non-sequential models produce similarly interpretable clusters -- thus, we use the \texttt{BoE+Attn} model. 
Similar to \citet{Iyyer2016}, we limit the input vocabulary for both \texttt{BoW + Attn} encoders to words occurring in at least 50 movies (7.3\% of the training set), outside the 500 most frequent words. 

The number of descriptors $k$ is set to 25 to allow for a wide range of topics while keeping manual examination feasible.
Descriptors are initialized either randomly \citep{Glorot2010} or with the centroids of a $k$-means clustering of the input word embeddings.
For the predictor, $f$ is a two-layer FFNN with ReLU activations and a softmax final layer that transforms $\mathbf{v}_t$ (from the scene encoder) into a 100-dimensional intermediate state and then into $\mathbf{o}_t$.
Further modeling choices are evaluated using the semantic coherence metric \citep{Mimno2011}, which assesses the quality of word clusters induced by topic modeling algorithms. 
These choices include: the presence of recurrence in the predictor (i.e., toggling between Eqns. \ref{eq:predictor} and \ref{eq:recurrent_predictor}, with $\alpha=0.5$) and the value of hyperparameter $\lambda$.
While the $k$-means initialized descriptors score slightly higher on semantic coherence, they are qualitatively quite similar to the initial centroids and do not reflect the corpus as well as the randomly initialized version.
We also find that incorporating recurrence and $\lambda=10$ (tuned using simple grid search) result in the highest coherence.

The outputs of the scene descriptor model are shown in Table \ref{tab:clusters} and Figure \ref{fig:descriptor_trajectories}.
Table \ref{tab:clusters} presents five example descriptors, each identified by representative words closest to them in the word embedding space, with their topic names manually annotated.
Figure \ref{fig:descriptor_trajectories} presents the corresponding narrative trajectories of a subset of these descriptors over the course of three sample screenplays: Pretty Woman, Pulp Fiction, and Pearl Harbor, using a \emph{streamgraph} \citep{Byron2008}. 
The descriptor weight $\mathbf{o}_t$ (Eq.\ref{eq:predictor}) as a function of scene order is rescaled and smoothed, with the width of a region at a given scene indicating the weight value.
A critical event for each screenplay is indicated by a letter on each trajectory.
A qualitative analysis of such events indicates general alignment between scripts and their topic trajectories, and the potential applicability of this method to identifying significant moments in long-form documents.

\begin{table}[h!]
\small
\vspace{-0.2cm}
\begin{center}
\begin{tabular}{p{40pt}|l}
\textbf{Topic} & \textbf{Words} \\
\hline
Violence & fires blazes explosions grenade blasts \\
Residential & loft terrace courtyard foyer apartments \\
Military & leadership army victorious commanding elected \\
Vehicles & suv automobile wagon sedan cars \\
Geography & sand slope winds sloping cliffs
\end{tabular}
\end{center}
\caption{\footnotesize Examples of retrieved descriptors. Trajectories for ``Violence", ``Military", and ``Residential" are shown in Fig. \ref{fig:descriptor_trajectories}.}
\label{tab:clusters}
\vspace{-0.2cm}
\end{table}

\section{Related Work}
\label{sec:related}
Computational narrative analysis of large texts has been explored in a number of contexts \citep{Mani2012} and for a number of years \citep{Lehnert1981}.
More recent work has analyzed narrative from a plot \citep{Chambers2008,Goyal2010} and character \citep{Elsner2012,Bamman2014} perspective.
While movie narratives have received attention \citep{Bamman2013,Chaturvedi2018,Kar2018}, the computational analysis of entire screenplays has not been as common.

Notably, \citet{Gorinski2015} introduced a summarization method that takes into account an entire script at a time, extracting graph-based features that summarize the key scene sequences.
\citet{Gorinski2018} then build on top of this work, crafting additional features for use in a specially-designed multi-label encoder.
Our work suggests an orthogonal approach -- our automatically learned scene representations offer an alternative to their feature-engineered inputs.

\citet{Gorinski2018} emphasize the difficulty of their tag prediction task, which we find in our tasks as well.
One possibility we consider is that at least some of this difficulty owes not to the length or richness of the text per se, but rather to the complexity of the tag taxonomy.
The pattern of results we obtain from a similarity-based scoring measure offers a significantly brighter picture of model performance, and suggests more broadly that the standard multilabel F1 measure may not be appropriate for complex, human-crafted tag sets \citep{Maynard2006}.

Nevertheless, dealing with long-form text remains a significant challenge.
One possible solution is to infer richer representations of latent structure by using a structured attention mechanism \citep{Liu2018}, which might highlight key dependencies between scenes in a script.
Another method could be to define auxiliary tasks as in \citet{Jiang2018} to encourage better selection and memorization.
Lastly, sparse versions of the softmax function \citep{Martins2016} can be used to enforce the notion that salient information for downstream tasks is sparsely distributed across the screenplay. 

\section{Conclusion}
In this work, we propose and evaluate various neural network architectures for learning fixed-dimensional representations of full-length film scripts.
We hypothesize that designing the network to mimic the documents' internal structure will boost performance. Experiments conducted on two tag prediction tasks provide evidence in favour of this hypothesis, confirming the benefits of (1) using hierarchical attention-based models and (2) incorporating distinctions between different kinds of scene components directly into the model.
Additionally, as a means of exploring the information contained within scene-level embeddings, we presented an unsupervised technique for bootstrapping ``scene descriptors" and visualizing their trajectories through the screenplay.

For future work, we plan to investigate richer ways of incorporating character identities into the model. 
For example, character embeddings could be used to analyze character archetypes across different movies.
A persona-based characterization of the screenplay would provide a complementary view to the plot-based analysis elucidated here.

Finally, as noted at the outset, our structure-aware methods are fundamentally generalizable, and can be adapted to natural language understanding across virtually any domain in which structure can be extracted, including books, technical reports, and online chat logs, among others.

\bibliography{emnlp2020}

\begin{thebibliography}{46}
\expandafter\ifx\csname natexlab\endcsname\relax\def\natexlab#1{#1}\fi

\bibitem[{Adi et~al.(2016)Adi, Kermany, Belinkov, Lavi, and Goldberg}]{Adi2016}
Yossi Adi, Einat Kermany, Yonatan Belinkov, Ofer Lavi, and Yoav Goldberg. 2016.
\newblock Fine-grained analysis of sentence embeddings using auxiliary
  prediction tasks.
\newblock \emph{Proceedings of ICLR}.

\bibitem[{Argentini(1998)}]{Argentini1998}
Paul Argentini. 1998.
\newblock \emph{Elements of Style for Screenwriters}.
\newblock Lone Eagle Publishing.

\bibitem[{Bahdanau et~al.(2015)Bahdanau, Cho, and Bengio}]{Bahdanau2015}
Dzmitry Bahdanau, Kyunghyun Cho, and Yoshua Bengio. 2015.
\newblock Neural machine translation by jointly learning to align and
  translate.
\newblock In \emph{Proceedings of ICLR}.

\bibitem[{Bamman et~al.(2013)Bamman, O'Connor, and Smith}]{Bamman2013}
David Bamman, Brendan O'Connor, and Noah~A. Smith. 2013.
\newblock Learning latent personas of film characters.
\newblock In \emph{Proceedings of ACL}.

\bibitem[{Bamman et~al.(2014)Bamman, Underwood, and Smith}]{Bamman2014}
David Bamman, Ted Underwood, and Noah~A. Smith. 2014.
\newblock A bayesian mixed effects model of literary character.
\newblock In \emph{Proceedings of ACL}.

\bibitem[{Barrault et~al.(2019)Barrault, Bojar, Costa-juss{\`a}, Federmann,
  Fishel, Graham, Haddow, Huck, Koehn, Malmasi, Monz, M{\"u}ller, Pal, Post,
  and Zampieri}]{Barrault2019}
Lo{\"\i}c Barrault, Ond{\v{r}}ej Bojar, Marta~R. Costa-juss{\`a}, Christian
  Federmann, Mark Fishel, Yvette Graham, Barry Haddow, Matthias Huck, Philipp
  Koehn, Shervin Malmasi, Christof Monz, Mathias M{\"u}ller, Santanu Pal, Matt
  Post, and Marcos Zampieri. 2019.
\newblock Findings of the 2019 conference on machine translation (wmt19).
\newblock In \emph{Proceedings of WMT}.

\bibitem[{Bhagavatula et~al.(2018)Bhagavatula, Feldman, Power, and
  Ammar}]{Bhagavatula2018}
Chandra Bhagavatula, Sergey Feldman, Russell Power, and Waleed Ammar. 2018.
\newblock Content-based citation recommendation.
\newblock In \emph{Proceedings NAACL}.

\bibitem[{Blei and Lafferty(2006)}]{Blei2006}
David~M. Blei and John~D. Lafferty. 2006.
\newblock Dynamic topic models.
\newblock In \emph{Proceedings of ICML}.

\bibitem[{Byron and Wattenberg(2008)}]{Byron2008}
L~Byron and M.~Wattenberg. 2008.
\newblock Stacked graphs – geometry aesthetics.
\newblock \emph{IEEE Transactions on Visualization and Computer Graphics}.

\bibitem[{Chambers and Jurafsky(2008)}]{Chambers2008}
Nathanael Chambers and Dan Jurafsky. 2008.
\newblock Unsupervised learning of narrative event chains.
\newblock In \emph{Proceedings of ACL}.

\bibitem[{Chaturvedi et~al.(2018)Chaturvedi, Srivastava, and
  Roth}]{Chaturvedi2018}
Snigdha Chaturvedi, Shashank Srivastava, and Dan Roth. 2018.
\newblock Where have {I} heard this story before? identifying narrative
  similarity in movie remakes.
\newblock In \emph{Proceedings of NAACL}.

\bibitem[{Conneau et~al.(2018)Conneau, Kruszewski, Lample, Barrault, and
  Baroni}]{Conneau2018}
Alexis Conneau, German Kruszewski, Guillaume Lample, Lo{\"\i}c Barrault, and
  Marco Baroni. 2018.
\newblock What you can cram into a single vector: Probing sentence embeddings
  for linguistic properties.
\newblock \emph{Proceedings of ACL}.

\bibitem[{Elsner(2012)}]{Elsner2012}
Micha Elsner. 2012.
\newblock Character-based kernels for novelistic plot structure.
\newblock In \emph{Proceedings of EACL}.

\bibitem[{Gardner et~al.(2017)Gardner, Grus, Neumann, Tafjord, Dasigi, Liu,
  Peters, Schmitz, and Zettlemoyer}]{Gardner2017}
Matt Gardner, Joel Grus, Mark Neumann, Oyvind Tafjord, Pradeep Dasigi,
  Nelson~F. Liu, Matthew Peters, Michael Schmitz, and Luke~S. Zettlemoyer.
  2017.
\newblock \href {http://arxiv.org/abs/arXiv:1803.07640} {Allennlp: A deep
  semantic natural language processing platform}.

\bibitem[{Glorot and Bengio(2010)}]{Glorot2010}
Xavier Glorot and Yoshua Bengio. 2010.
\newblock Understanding the difficulty of training deep feedforward neural
  networks.
\newblock In \emph{Procedings of AIStats}.

\bibitem[{Gorinski and Lapata(2015)}]{Gorinski2015}
Philip~John Gorinski and Mirella Lapata. 2015.
\newblock Movie script summarization as graph-based scene extraction.
\newblock In \emph{Proceedings of NAACL}.

\bibitem[{Gorinski and Lapata(2018)}]{Gorinski2018}
Philip~John Gorinski and Mirella Lapata. 2018.
\newblock What{'}s this movie about? a joint neural network architecture for
  movie content analysis.
\newblock In \emph{Proceedings of NAACL}.

\bibitem[{Goyal et~al.(2010)Goyal, Riloff, and Daum{\'e}}]{Goyal2010}
Amit Goyal, Ellen Riloff, and Hal Daum{\'e}, III. 2010.
\newblock Automatically producing plot unit representations for narrative text.
\newblock In \emph{Proceedings of EMNLP}.

\bibitem[{He et~al.(2017)He, Lee, Ng, and Dahlmeier}]{He2017unsupervised}
Ruidan He, Wee~Sun Lee, Hwee~Tou Ng, and Daniel Dahlmeier. 2017.
\newblock An unsupervised neural attention model for aspect extraction.
\newblock In \emph{Proceedings of ACL}.

\bibitem[{Iyyer et~al.(2016)Iyyer, Guha, Chaturvedi, Boyd-Graber, and
  Daum{\'e}~III}]{Iyyer2016}
Mohit Iyyer, Anupam Guha, Snigdha Chaturvedi, Jordan Boyd-Graber, and Hal
  Daum{\'e}~III. 2016.
\newblock Feuding families and former {F}riends: Unsupervised learning for
  dynamic fictional relationships.
\newblock In \emph{Proceedings of NAACL}.

\bibitem[{Ji and Smith(2017)}]{Ji2017}
Yangfeng Ji and Noah~A. Smith. 2017.
\newblock Neural discourse structure for text categorization.
\newblock In \emph{Proceedings of ACL}.

\bibitem[{Jiang and Bansal(2018)}]{Jiang2018}
Yichen Jiang and Mohit Bansal. 2018.
\newblock Closed-book training to improve summarization encoder memory.
\newblock In \emph{Proceedings of EMNLP}.

\bibitem[{Kar et~al.(2018)Kar, Maharjan, and Solorio}]{Kar2018}
Sudipta Kar, Suraj Maharjan, and Thamar Solorio. 2018.
\newblock {F}olksonomication: Predicting tags for movies from plot synopses
  using emotion flow encoded neural network.
\newblock In \emph{Proceddings of COLING}.

\bibitem[{Ko{\v{c}}isk{\'y} et~al.(2018)Ko{\v{c}}isk{\'y}, Schwarz, Blunsom,
  Dyer, Hermann, Melis, and Grefenstette}]{Kocisky2018}
Tom{\'a}{\v{s}} Ko{\v{c}}isk{\'y}, Jonathan Schwarz, Phil Blunsom, Chris Dyer,
  Karl~Moritz Hermann, G{\'a}bor Melis, and Edward Grefenstette. 2018.
\newblock The {N}arrative{QA} reading comprehension challenge.
\newblock \emph{Transactions of the Association for Computational Linguistics},
  6.

\bibitem[{Lehnert(1981)}]{Lehnert1981}
Wendy~G. Lehnert. 1981.
\newblock Plot units and narrative summarization.
\newblock \emph{Cognitive Science}, 5(4).

\bibitem[{Linzen et~al.(2016)Linzen, Dupoux, and Goldberg}]{Linzen2016}
Tal Linzen, Emmanuel Dupoux, and Yoav Goldberg. 2016.
\newblock Assessing the ability of {LSTM}s to learn syntax-sensitive
  dependencies.
\newblock \emph{Transactions of the ACL}.

\bibitem[{Liu and Lapata(2018)}]{Liu2018}
Yang Liu and Mirella Lapata. 2018.
\newblock Learning structured text representations.
\newblock \emph{Transactions of the Association for Computational Linguistics},
  6.

\bibitem[{Mani(2012)}]{Mani2012}
Inderjeet Mani. 2012.
\newblock \emph{Synthesis Lectures on Human Language Technologies:
  Computational Modeling of Narrative}.
\newblock Morgan Claypool.

\bibitem[{Martins and Astudillo(2016)}]{Martins2016}
Andre Martins and Ramon Astudillo. 2016.
\newblock From softmax to sparsemax: A sparse model of attention and
  multi-label classification.
\newblock In \emph{Proceedings ICML}.

\bibitem[{Maynard et~al.(2006)Maynard, Peters, and Li}]{Maynard2006}
Diana Maynard, Wim Peters, and Yaoyong Li. 2006.
\newblock Metrics for evaluation of ontology-based information extraction.
\newblock In \emph{CEUR Workshop Proceedings}.

\bibitem[{Mimno et~al.(2011)Mimno, Wallach, Talley, Leenders, and
  McCallum}]{Mimno2011}
David Mimno, Hanna~M Wallach, Edmund Talley, Miriam Leenders, and Andrew
  McCallum. 2011.
\newblock Optimizing semantic coherence in topic models.
\newblock In \emph{Proceedings of EMNLP}.

\bibitem[{Mitcheltree et~al.(2018)Mitcheltree, Wharton, and
  Saluja}]{Mitcheltree2018}
Christopher Mitcheltree, Skyler Wharton, and Avneesh Saluja. 2018.
\newblock Using aspect extraction approaches to generate review summaries and
  user profiles.
\newblock In \emph{Proceedings of NAACL}.

\bibitem[{{Nguyen}(2017)}]{Nguyen2017}
Dat~Quoc {Nguyen}. 2017.
\newblock {An overview of embedding models of entities and relationships for
  knowledge base completion}.
\newblock \emph{arXiv e-prints}.

\bibitem[{Olshausen and Field(1997)}]{Olshausen1997}
Bruno~A Olshausen and David~J Field. 1997.
\newblock Sparse coding with an overcomplete basis set: A strategy employed by
  v1?
\newblock \emph{Vision Research}.

\bibitem[{Paszke et~al.(2019)Paszke, Gross, Massa, Lerer, Bradbury, Chanan,
  Killeen, Lin, Gimelshein, Antiga, Desmaison, Kopf, Yang, DeVito, Raison,
  Tejani, Chilamkurthy, Steiner, Fang, Bai, and Chintala}]{Paszke2019}
Adam Paszke, Sam Gross, Francisco Massa, Adam Lerer, James Bradbury, Gregory
  Chanan, Trevor Killeen, Zeming Lin, Natalia Gimelshein, Luca Antiga, Alban
  Desmaison, Andreas Kopf, Edward Yang, Zachary DeVito, Martin Raison, Alykhan
  Tejani, Sasank Chilamkurthy, Benoit Steiner, Lu~Fang, Junjie Bai, and Soumith
  Chintala. 2019.
\newblock Pytorch: An imperative style, high-performance deep learning library.
\newblock In \emph{Proceedings of NeurIPS}.

\bibitem[{Pavel et~al.(2015)Pavel, Goldman, Hartmann, and Agrawala}]{Pavel2015}
Amy Pavel, Dan~B. Goldman, Bj\"{o}rn Hartmann, and Maneesh Agrawala. 2015.
\newblock Sceneskim: Searching and browsing movies using synchronized captions,
  scripts and plot summaries.
\newblock In \emph{Proceedings of UIST}.

\bibitem[{Pennington et~al.(2014)Pennington, Socher, and
  Manning}]{Pennington2014}
Jeffrey Pennington, Richard Socher, and Christopher~D. Manning. 2014.
\newblock Glove: Global vectors for word representation.
\newblock In \emph{Proceedings of EMNLP}.

\bibitem[{Ren et~al.(2014)Ren, Liu, Yu, Khandelwal, Gu, Wang, and
  Han}]{Ren2014}
Xiang Ren, Jialu Liu, Xiao Yu, Urvashi Khandelwal, Quanquan Gu, Lidan Wang, and
  Jiawei Han. 2014.
\newblock Cluscite: Effective citation recommendation by information
  network-based clustering.
\newblock In \emph{Proceedings of KDD}.

\bibitem[{Shi et~al.(2016)Shi, Padhi, and Knight}]{Shi2016}
Xing Shi, Inkit Padhi, and Kevin Knight. 2016.
\newblock Does string-based neural mt learn source syntax?
\newblock In \emph{Proceedings of EMNLP}.

\bibitem[{Socher et~al.(2013)Socher, Perelygin, Wu, Chuang, Manning, Ng, and
  Potts}]{Socher2013}
Richard Socher, Alex Perelygin, Jean Wu, Jason Chuang, Christopher~D. Manning,
  Andrew Ng, and Christopher Potts. 2013.
\newblock Recursive deep models for semantic compositionality over a sentiment
  treebank.
\newblock In \emph{Proceedings of EMNLP}.

\bibitem[{Strubell et~al.(2018)Strubell, Verga, Andor, Weiss, and
  McCallum}]{Strubell2018}
Emma Strubell, Patrick Verga, Daniel Andor, David Weiss, and Andrew McCallum.
  2018.
\newblock Linguistically-informed self-attention for semantic role labeling.
\newblock In \emph{Proceedings of EMNLP}.

\bibitem[{Sukhbaatar et~al.(2015)Sukhbaatar, szlam, Weston, and
  Fergus}]{Sukhbaatar2015}
Sainbayar Sukhbaatar, arthur szlam, Jason Weston, and Rob Fergus. 2015.
\newblock End-to-end memory networks.
\newblock In \emph{Proceedings of NeurIPS}.

\bibitem[{Wang et~al.(2019)Wang, Singh, Michael, Hill, Levy, and
  Bowman}]{Wang2019}
Alex Wang, Amanpreet Singh, Julian Michael, Felix Hill, Omer Levy, and
  Samuel~R. Bowman. 2019.
\newblock {GLUE}: A multi-task benchmark and analysis platform for natural
  language understanding.
\newblock In \emph{Proceedings of ICLR}.

\bibitem[{Weston et~al.(2011)Weston, Bengio, and Usunier}]{Weston2011}
Jason Weston, Samy Bengio, and Nicolas Usunier. 2011.
\newblock Wsabie: Scaling up to large vocabulary image annotation.
\newblock In \emph{Proceedings of IJCAI}.

\bibitem[{Williams et~al.(2018)Williams, Nangia, and Bowman}]{Williams2018}
Adina Williams, Nikita Nangia, and Samuel Bowman. 2018.
\newblock A broad-coverage challenge corpus for sentence understanding through
  inference.
\newblock In \emph{Proceedings of NAACL}.

\bibitem[{Yang et~al.(2016)Yang, Yang, Dyer, He, Smola, and Hovy}]{Yang2016}
Zichao Yang, Diyi Yang, Chris Dyer, Xiaodong He, Alex Smola, and Eduard Hovy.
  2016.
\newblock Hierarchical attention networks for document classification.
\newblock In \emph{Proceedings of NAACL}.

\end{thebibliography}
\bibliographystyle{acl_natbib}

\pagebreak
\appendix
\section{Appendix}
\subsection{Additional Dataset Statistics}
\label{sec:add_stats}
In this section, we present additional statistics on the evaluation sets used in this work. 

\begin{table}[!h]
\begin{tabular}{cccc}
\textbf{Min} & \textbf{10th \%} & \textbf{90th \%} & \textbf{Max} \\
\hline
4025 & 16,240 & 29,376 & 52,059
\end{tabular}
\caption{Statistics on the number of tokens per script in the Scriptbase-J corpus. We use the same script corpus with two different tag sets -- the Jinni tags provided with ScriptBase and a tag set designed by internal annotators.}
\label{tab:token_stats}
\end{table}

\begin{table}[!h]
\small
\begin{tabular}{l|ccc}
\multicolumn{1}{l|}{\textbf{Tag}} & \multicolumn{1}{c}{\textbf{\begin{tabular}[c]{@{}c@{}}Value\end{tabular}}} \\
\hline
\textbf{Genre} & Crime, Independent \\
\textbf{Mood} & Clever, Witty, Stylized \\
\textbf{Attitude} & Semi Serious, Realistic \\
\textbf{Plot} & Tough Heroes, Violence Spree, On the Run \\ 
\textbf{Place} & California, Los Angeles, Urban \\
\textbf{Flag} & Drugs/Alcohol, Profanity, Violent Content \\ 
\hline
\textbf{Logline} & ``The lives of two mob hit men, a boxer, \\
& a gangster's wife, and a pair of diner \\
& bandits intertwine in four tales of \\
& violence and redemption."
\end{tabular}
\caption{Examples of Scriptbase-J tag attributes, tag values, and a logline, for  the film ``Pulp Fiction".}
\label{tab:tag_examples}
\end{table}

\begin{table}[!h]
\begin{tabular}{l|ccc}
\multicolumn{1}{l|}{\textbf{Tag}} & \multicolumn{1}{c}{\textbf{\begin{tabular}[c]{@{}c@{}}Internal\end{tabular}}} & \multicolumn{1}{c}{\textbf{\begin{tabular}[c]{@{}c@{}}Scriptbase-J\end{tabular}}} \\
\hline
Genre & 9 & 31 \\
Mood & 65 & 18 \\
Attitude & - & 8 \\
Plot & 82 & 101 \\
Place & - & 24 \\
Flag & - & 6 \\
\end{tabular}
\caption{The number of distinct tag values for each tag attribute across the two datasets. Cardinalities for Scriptbase-J tag attributes are identical to \citet{Gorinski2018} except for the removal of one mood tag value when filtering for erroneously preprocessed scripts.}
\label{tab:tag_cardinality}
\end{table}

\begin{table}[!h]
\small
\begin{tabular}{l|ccc}
\multicolumn{1}{l|}{\textbf{Tag}} & \multicolumn{1}{c}{\textbf{\begin{tabular}[c]{@{}c@{}}Avg.\\ \#tags/script\end{tabular}}} & \multicolumn{1}{c}{\textbf{\begin{tabular}[c]{@{}c@{}}Min\\ \#scripts/tag\end{tabular}}} & \multicolumn{1}{c}{\textbf{\begin{tabular}[c]{@{}c@{}}Max\\ \#scripts/tag\end{tabular}}} \\
\hline
Genre & 1.74 & 17 & 347 \\
Mood & 3.29 & 15 & 200 \\
Plot & 2.50 & 15 & 73
\end{tabular}
\caption{Statistics for the three tag attributes applied in our internally-tagged dataset: average number of tags per script, and the minimum/maximum number of movies associated with any single value.}
\label{tab:tag_stats}
\end{table}

\subsection{Tag Similarity Scoring}

\begin{table}[!h]
\begin{tabular}{l|ccc}
\multicolumn{1}{l|}{\textbf{Tag}} & \multicolumn{1}{c}{\textbf{\begin{tabular}[c]{@{}c@{}}Target\end{tabular}}} &
\multicolumn{1}{c}{\textbf{\begin{tabular}[c]{@{}c@{}}Similar\end{tabular}}} &
\multicolumn{1}{c}{\textbf{\begin{tabular}[c]{@{}c@{}}Unrelated\end{tabular}}}\\
\hline
\textbf{Genre} & Period & Historical & Fantasy \\
\textbf{Mood} & Witty & Humorous & Bleak  \\
\textbf{Plot} & Hitman & Deadly & Love/Romance  \\ 
\end{tabular}
\caption{Examples of closely related and unrelated tag values in the Scriptbase-J tag set.}
\label{tab:tag_similarity}
\end{table}

To estimate tag-tag similarity percentiles, we calculate the distance between tag embeddings learned via an auxiliary model trained on a related supervised learning task.
In our case, the related task is to predict the audience segment of a movie, given a tag set. 
The general approach is easily replicable via any model that projects tags into a well-defined similarity space (e.g., knowledge-graph embeddings \citep{Nguyen2017} or tag-based autoencoders).

Given a tag embedding space, the similarity percentile of a pair of tag values is estimated as follows.
For a given tag attribute, the pairwise cosine distance between tag embeddings is computed for all tag-tag value pairs.
For a given pair, its similarity percentile is then calculated with reference to the overall distribution for that attribute.

Similarity thresholding simplifies the tag prediction task by significantly reducing the \emph{perplexity} of the tag set, while only marginally reducing its \emph{cardinality}. 
Cardinality can be estimated via permutations. If $n$ is the cardinality of the tag set, the number of permutations $p$ of different tag pairs ($k=2$) is:
\begin{align}
p(n,k) = \frac{n!}{(n-k)!}  
\label{eq:permutations}
\end{align}
which simplifies to $n^{2}-n-p=0$.

Likewise, the entropy of a list of $n$ distinct tag values of varying probabilities is given by:
\begin{align}H(X) = H(\text{tag}_{1},...,\text{tag}_{n}) = -\sum^{n}_{i=1}\text{tag}_{i}\log_{2}\text{tag}_{i}
\label{eq:entropy}
\end{align}
The perplexity over tags is then simply $2^{H(X)}$. 

\begin{table}[!h]
\begin{tabular}{l|ccc}
\multicolumn{1}{l|}{\textbf{Tag}} & \multicolumn{1}{c}{\textbf{\begin{tabular}[c]{@{}c@{}}Perplexity\end{tabular}}} & \multicolumn{1}{c}{\textbf{\begin{tabular}[c]{@{}c@{}}Cardinality\end{tabular}}} \\
\hline
Genre & 42\% & 16\% \\
Mood & 77\% & 16\% \\
Plot & 79\% & 16\% \\
\end{tabular}
\caption{The percent decrease in perplexity and cardinality, respectively, as the similarity threshold decreases from 100th percentile similarity (baseline) to 70th percentile.}
\label{tab:perplexity_cardinality}
\end{table}

As the similarity threshold decreases, the number of tags treated as equivalent correspondingly increases. 
Mapping these ``equivalents" to a shared label in our list of tag values allows us to calculate updated values for tag (1) perplexity and (2) cardinality.
As illustrated by Table \ref{tab:perplexity_cardinality}, rather than leading to large reductions in the overall cardinality of the tag set, similarity thresholding mainly serves to decrease perplexity by eliminating redundant/highly similar alternatives. 
Thus, thresholding at once significantly decreases the complexity of the prediction task, while yielding a potentially more representative picture of model performance.

\end{document}